\documentclass[letterpaper, 10 pt, journal, twoside]{IEEEtran}
\usepackage{amsmath,amsfonts}
\usepackage{algorithmic}
\usepackage{algorithm}
\usepackage{array}
\usepackage[caption=false,font=normalsize,labelfont=sf,textfont=sf]{subfig}
\usepackage{textcomp}
\usepackage{stfloats}
\usepackage{url}
\usepackage{verbatim}
\usepackage{graphicx}
\usepackage[colorlinks=true, urlcolor=blue, linkcolor=red]{hyperref}
\usepackage{cite}
\usepackage[]{multirow}

\newcommand{\bfx}{\boldsymbol{x}}
\newcommand{\bfz}{\boldsymbol{z}}

\newcommand{\bfy}{\boldsymbol{y}}

\begin{document}

\title{Knowledge-based Neural Ordinary Differential Equations for Cosserat Rod-based Soft Robots}

\author{Tom Z. Jiahao, Ryan Adolf, Cynthia Sung, and M. Ani Hsieh
\thanks{We gratefully acknowledge the support of NSF DCSD-2121887.}
\thanks{GRASP Laboratory, University of Pennsylvania, Philadelphia, USA.
        {\tt\footnotesize \{zjh, ryanaa, crsung, m.hsieh\}@seas.upenn.edu}}

\thanks{Preprint revised August 8, 2024.}}


\IEEEaftertitletext{\vspace{-2\baselineskip}} 
\maketitle

\begin{abstract}
Soft robots have many advantages over rigid robots thanks to their compliant and passive nature. However, it is generally challenging to model the dynamics of soft robots due to their high spatial dimensionality, making it difficult to use model-based methods to accurately control soft robots. It often requires direct numerical simulation of partial differential equations to simulate soft robots. This not only requires an accurate numerical model, but also makes soft robot modeling slow and expensive. Deep learning algorithms have shown promises in data-driven modeling of soft robots. However, these algorithms usually require a large amount of data, which are difficult to obtain in either simulation or real-world experiments of soft robots. In this work, we propose KNODE-Cosserat, a framework that combines first-principle physics models and neural ordinary differential equations. We leverage the best from both worlds -- the generalization ability of physics-based models and the fast speed of deep learning methods. We validate our framework in both simulation and real-world experiments. In both cases, we show that the robot model significantly improves over the baseline models under different metrics. 
\end{abstract}


\section{Introduction}

The domain of soft robotics lies at the intersection of material science, mechanical engineering, control systems, and artificial intelligence. Unlike their rigid counterparts, soft robots offer better versatility, adaptability, and safety for human interaction. However, modeling the dynamics of soft robots poses significant challenges due to their inherent nonlinear and high-dimensional characteristics~\cite{Rus2015, doi:10.1126/scirobotics.aah3690, trivedi}. Traditional modeling techniques for rigid robotics, often grounded in first-principle physics, struggle to encapsulate the complex behaviors of soft robots, especially under varying and unpredictable environmental conditions~\cite{Rus2015, doi:10.1126/scirobotics.aah3690, trivedi}.

The advent of machine learning, particularly neural networks, has opened new possibilities through data-driven modeling and control of soft robots. Neural networks excel in identifying patterns and learning from data, enabling them to capture complex dynamics that evade conventional modeling frameworks. However, while neural networks offer impressive predictive capabilities, they often lack the interpretability and generalizability that first-principle physics-based models provide. Our work seeks to bridge this gap by proposing a hybrid modeling approach that combines neural networks with first-principle physics models. This integration aims to harness the robustness and interpretability of physics-based models along with the flexibility and adaptability of neural networks. a

\begin{figure}[H]
\begin{center}\includegraphics[width=0.98\columnwidth]{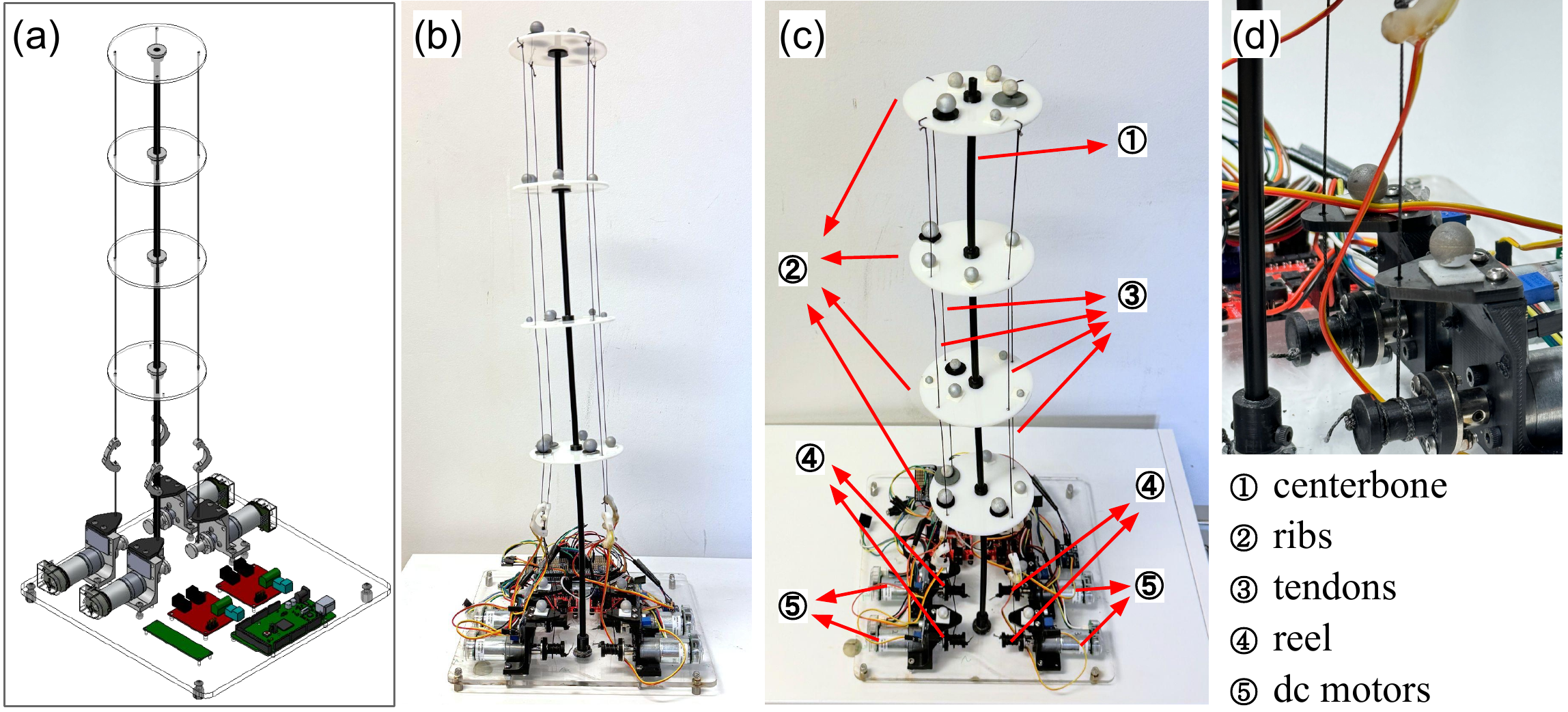}
    \end{center}
    \caption{\textbf{Custom built open-source tendon-driven continuum robot.} (a) CAD model of the robot. (b) Front view of the robot. (c) Isometric view with components annotations. (d) Close-up of the reel mechanism.}
    \label{fig: custom robot annotated}
\end{figure}

We adopt knowledge-based neural ordinary differential equations (KNODE) to model the dynamics of soft robots. In previous works, NODE usually uses neural networks to model temporal derivatives. However in this work, we use NODE to improve the accuracy of spatial derivatives. We apply our method on a general class of soft robot systems based on the Cosserat rod model.

The two primary contributions of this paper are as follows:

\begin{itemize}
    \item We introduce a novel hybrid modeling framework for soft robotics that integrates the reliability of physics-based models with the adaptability of neural networks. This framework leverages the strengths of both approaches to provide accurate modeling of soft robot dynamics.

    \item We demonstrate using knowledge-based neural ordinary differential equations in spatial dynamics. To the authors' knowledge, this is the first time any NODE variant is applied to continuous-space systems for robotics applications. 

\end{itemize}    


\section{Related Works}
Physics-based models serve as the cornerstone for understanding and predicting the behavior of soft robots. These models utilize principles from continuum mechanics to describe the dynamics and deformation of soft structures. One prominent approach is the Cosserat rod theory, which extends traditional beam theories to capture the bending, twisting, stretching, and shearing of slender structures~\cite{till2019real, rucker2011statics, QinReport}. The Euler-Bernoulli beam theory, for instance, has been effectively applied to model the bending deformations of soft robotic actuators~\cite{Bauchau2009, MbakopInverseDynamics}. Despite their interpretability and generalizability, physics-based models often require significant effort to derive and fine-tune. Additionally, these models may struggle with complex physics, such as contact and anisotropic material properties, which are prevalent in real-world applications~\cite{Webster2010DesignAK, Bergou2008DiscreteER}.

Finite Element Analysis (FEA) is extensively used for the detailed simulation of soft robots due to its ability to model complex geometries and material properties with high fidelity. FEA subdivides the robotic structure into smaller, finite elements and solves the governing equations for each element, providing a comprehensive understanding of the internal and external forces at play~\cite{zhang:hal-01370347, Guo2020SimulationAF}. This method excels in capturing the detailed mechanical behavior of soft robots, including intricate deformations and interactions with the environment~\cite{ChenClimb, Goury2018FastGA}. However, the high computational cost associated with FEA poses a challenge for its use in real-time control scenarios. The detailed nature of FEA simulations often requires significant computational resources and time, making them less suitable for applications requiring rapid responses~\cite{katzschmann2019dynamically, li2022equivalent}.

Data-driven approaches, including Physics-Informed Neural Networks (PINNs), Koopman operators, and other machine learning methods, have emerged as powerful tools for modeling soft robots. These methods leverage large datasets to train models that can predict the behavior of soft robots under various conditions. PINNs, for example, incorporate physical laws into the learning process, ensuring that predictions adhere to fundamental principles~\cite{RAISSI2019686, zhu2019physics}. Koopman operators express a system's dynamics as linear models in an infinite-dimensional space, enabling easier analysis and reliable controller design~\cite{Shi2023, shi2024koopman}. Other machine learning approaches, such as deep reinforcement learning and neural network-based models, have been used to capture complex dynamics and control strategies~\cite{9369003, 9561145, Truby2020DistributedPO}. The primary advantage of data-driven models lies in their adaptability and speed. Once trained, these models can provide rapid predictions, making them suitable for real-time applications. However, the need for extensive datasets and the potential lack of interpretability remain significant challenges. Obtaining sufficient training data can be difficult and expensive, and the black-box nature of many machine learning models can make it challenging to understand and trust their predictions~\cite{fang2022efficient, Truby2020DistributedPO}.

The recent trend with data-driven modeling has been towards hybrid approaches, which attempt to combine the generalizability of physics-based models with the adaptiveness of neural networks. Notably, knowledge-based neural ordinary differential equations (KNODE) has demonstrated effectiveness at modeling robotic systems~\cite{Jiahao2021Knowledgebased, jiahao2022online, chee2022knode, 9811997}.

\section{Background}
\subsection{Generalized Cosserat-rod Dynamics}
In this work, we consider a general class of Cosserat-rod-based soft robot systems given by
\begin{equation}
\begin{aligned}
    A\bfy_s + B\bfy_{st} &= f_1,\\
    C\bfz + D\bfz_t &= f_2,
\end{aligned}
\label{eqn: general_eqn}
\end{equation}
where the vectors $\bfy$ and $\bfz$ are the state vectors describing a soft robot, and the subscripts $s$ and $t$ denotes their spatial and temporal derivatives respectively. The difference between $\bfy$ and $\bfz$ is that $\bfy$ is described by both temporal and spatial derivatives, while $\bfz$ is only described by its temporal derivatives. Furthermore, $A, B$ and $f_1$ are functions of $s, t, \bfy, \bfy_t, \bfz, \bfz_t$, and the control $\boldsymbol{\tau}$, while $C, D$, and $f_2$ are functions of $s, t, \bfy, \bfy_t$ and $\boldsymbol{\tau}$. In this paper, we denote the states at arclength $s$ and time $t$ as $\bfy(s, t)$ and $\bfz(s, t)$, and we use $\bfy(:, t)$ to denote \textit{all} spatial states at time $t$.

A wide class of soft robot systems can be described by \eqref{eqn: general_eqn}. For example, soft slender robot~\cite{xun2024cosserat}, variable curvature continuum robot~\cite{wang2024cosserat}, parallel continuum robot~\cite{till2019real}, and tendon-driven continuum robot~\cite{rucker2011statics}, can be modeled by equation \eqref{eqn: general_eqn}. To efficiently solve \eqref{eqn: general_eqn}, a method that combines implicit method and shooting method was proposed~\cite{till2019real}. This method semi-discretizes the spatial-temporal PDE into a system of equations that only involves the spatial derivatives using implicit method, which can then be solved as a boundary-value problem. For brevity, we will call the combined implicit method and shooting method as \textit{implicit-shooting method} in the rest of this paper.

\subsection{Implicit Method for Semi-discretization}
Backward differentiation formula (BDF), an implicit method, is applied to resolve the temporal derivatives in \eqref{eqn: general_eqn}. Assuming time step size $\delta t$, a temporal derivative can be approximated by
\begin{equation}
    \bfy_{t}(:, i) \approx c_0\bfy(:, i) + \sum^{\infty}_{k=1}[c_k\bfy(:,i-k) + d_k\bfy_t(:,i-k)],
\end{equation}
where the coefficients $c$ and $d$ are functions of the time step $\delta t$, and the temporal derivative of $\bfy$ at the $i$th step is approximated using the state of $\bfy$ at the current step and other terms which rely on the past history of the states. All the history terms can be lumped together to give the form
\begin{equation}
    \bfy_{t}(:,i) \approx c_0\bfy(:,i) + \overset{\text{h}}{\bfy}(:,i),
\end{equation}
where $\overset{\text{h}}{\bfy}(:, i)$ denotes the history terms for $\bfy_{t}$ at time $t=i$. Applying BDF, equation \eqref{eqn: general_eqn} becomes
\begin{equation}
    \begin{aligned}
        \bfy_s &= (A +c_0B)^{-1}\left(f_1 - B \overset{\text{h}}{\bfy}_s\right),\\
        \bfz &= (C + c_0D)^{-1}\left(f_2 - D \overset{\text{h}}{\bfz}\right).
    \end{aligned}
    \label{eq:semi-discretized general form}
\end{equation}
\subsection{Shooting Method for Boundary Value Problem}

The system \eqref{eq:semi-discretized general form} now only involves spatial derivatives, which can then be solved using shooting method as a boundary value problem. A solution is first guessed, and then the distal boundary conditions are optimized using iterative methods. The shooting method is summarized in Algorithm \ref{alg:shooting_method}, which returns the trajectory of the robot given its initial state. A common choice for the iterative method used to optimize for $\bfy(s)$ is Newton's method or its variants~\cite{197970}. In this paper, we utilize the \texttt{fsolve} function in the scipy library to optimize for zero boundary conditions.

\begin{algorithm}[ht]
\caption{Shooting method for boundary value problem}\label{alg:shooting_method}
\begin{algorithmic}
\STATE 
\STATE Initialize a robot with its initial state $\bfy(:, 0), \bfz(:, 0)$, control $\boldsymbol{\tau}(0)$, spatial discretization $S$ with spatial origin $s=0$, total simulation duration $T$ with starting time $t=0$, trajectory array $A = [\{\bfy(:, 0), \bfz(:, 0)\}]$, and known boundary conditions of the robot at $s=0$ and $S$. Initialize history terms $\overset{\text{h}}{\bfy}_s(:, 0)$ and $\overset{\text{h}}{\bfz}(:, 0)$ using only the initial states. 
\STATE $\textbf{While}$ $t < T$
\STATE \hspace{0.5cm}Guess the solution of the robot at $s=0$.
\STATE \hspace{0.5cm}Initialize the distal boundary residual to non-zero.

\STATE \hspace{0.5cm}$\textbf{While}$ distal boundary residual is not zero
\STATE \hspace{1cm}Step iterative solver for new solution at $s=0$.
\STATE \hspace{1cm}Integrate spatial ODE from $s=0$ to $S-1$.
\STATE \hspace{1cm}Evaluate distal boundary residual.
\STATE \hspace{0.5cm}$\textbf{EndWhile}$

\STATE \hspace{0.5cm}Append $\{\bfy(:, t),\bfz(: ,t)\}$ to $A$.
\STATE \hspace{0.5cm}Update control $\boldsymbol{\tau}(t+1)$.
\STATE \hspace{0.5cm}Update history terms $\overset{\text{h}}{\bfy}_s(:, t+1)$ and $\overset{\text{h}}{\bfz}(:, t+1)$.

\STATE $\textbf{EndWhile}$
\STATE \textbf{Return} $A$
\end{algorithmic}
\end{algorithm}

 \section{Problem Formulation}
While the equations \eqref{eq:semi-discretized general form} are generally effective in modeling the physics of Cosserat rod-based robots, it requires extensive tuning to match true physical hardware. Furthermore, the model makes many assumptions, such as uniform load across the centerbone and frictionless contact between the tendons and ribs in tendon-driven robots. These assumptions may fail in the real world. We seek to bridge the gap between the physics model and the true system using its observation data.

For more compact notations, we will write equation \eqref{eq:semi-discretized general form} as $\bfx = \mathbf{M}^{-1}\boldsymbol{f}$, where
\begin{equation}
\begin{aligned}
\bfx &= \begin{bmatrix} \bfy_s \\ \bfz \end{bmatrix},\\
\mathbf{f} &= \begin{bmatrix} f_1 - B\overset{\text{h}}{\bfy}_s \\ f_2 - D\overset{\text{h}}{\bfz} \end{bmatrix},\\
\mathbf{M} &= \begin{bmatrix} A + c_0 B & 0 \\ 0 & C + c_0 D \end{bmatrix}.
\end{aligned}
\end{equation}

Given a true Cosserat rod-based system and its trajectory $\mathcal{O} = [(\bfy(t_0),\bfz(t_0),\boldsymbol{\tau}(t_0)),$ $(\bfy(t_1),\bfz(t_1), \boldsymbol{\tau}(t_1)), \cdots,$ $(\bfy(t_N),\bfz(t_N), \boldsymbol{\tau}(t_N))]^T$, observed at the times $T = \{t_0, t_1,\cdots,t_N\}$. Also given an imperfect knowledge of the model $\bfx_{\text{im}} = \mathbf{M}_{\text{im}}^{-1}\boldsymbol{f}_{\text{im}}$ of the system, we assume that the underlying dynamics of the true system is given by
\begin{equation}
    \bfx = \mathbf{M}_{\text{im}}^{-1}\boldsymbol{f}_{\text{im}} + \Delta_{\boldsymbol{\theta}},
\end{equation}
where $\Delta_{\boldsymbol{\theta}}$ is the difference between the true model and the our knowledge, which is parametrized by $\boldsymbol{\theta}$. In this paper, we use a neural network to represent $\Delta_{\boldsymbol{\theta}}$, and $\boldsymbol{\theta}$ are its weights and biases.

We seek to model the difference $\Delta_{\boldsymbol{\theta}}$ such that $\mathbf{M}_{\text{im}}^{-1}\boldsymbol{f}_{\text{im}} + \Delta_{\boldsymbol{\theta}}$ is a good representation of the true underlying dynamics. Note that this is a more challenging problem than a parameter estimation problem in that the knowledge $\mathbf{M}_{\text{im}}$ and $\boldsymbol{f}_{\text{im}}$ could have different functional terms from the true system.

\section{Methodology}
\subsection{Knowledge-based Neural ODEs}
Neural ordinary differential equations (NODE), initially introduced to approximate continuous-depth residual networks~\cite{conf/nips/ChenRBD18}, have been subsequently employed in scientific machine learning to model a diverse array of dynamical systems~\cite{Jiahao2021Knowledgebased}. Knowledge-based neural ordinary differential equations (KNODE) extend NODE by integrating physics-based knowledge with neural networks, leveraging compatibility with fundamental principles to enhance the learning of continuous-time models with improved generalizability. While the original KNODE was applied solely to non-controlled dynamical systems, many robotic systems possess readily available physics models that can be utilized as knowledge. Recent variants have been developed to incorporate control inputs into dynamic models~\cite{chee2022knode, jiahao2022online}. In this work, we incorporate KNODE into learning continuous-space models. KNODE's suitability for our model learning task is underscored by its continuous nature and sample efficiency. To the authors' knowledge, this is the first time any NODE variant is applied to continuous-space systems for robotics applications. 

In the context of our learning task, we introduce a neural network $f_{\boldsymbol{\theta}}$ which models the residual of the spatial derivative, similar to~\cite{jiahao2022online, chee2022knode}. The overall dynamics can be written as the KNODE model given by
\begin{equation}
    \Tilde{\bfx} = \begin{bmatrix} \Tilde{\bfy}_s \\ \Tilde{\bfz} \end{bmatrix} = \mathbf{M}_{\text{im}}^{-1}\boldsymbol{f}_{\text{im}} + \boldsymbol{f}_{\boldsymbol{\theta}}(\bfy, \overset{\text{h}}{\bfy}_s,\bfz, \overset{\text{h}}{\bfz}, \boldsymbol{\tau}),
\label{eq:knode_dynamics}
\end{equation}
where $\boldsymbol{f}_{\boldsymbol{\theta}}$ is a neural network, and we use $\Tilde{\cdot}$ to denote predictions made by the KNODE model.

Given the initial condition $\bfx_0$ for a robot at $t=t_0$, its complete state at time $t=t_1$ can be obtained through the implicit-shooting method. In this work, we use a similar approach as~\cite{chee2022knode}, where the control is used during simulation steps but simply ignored when computing loss.

Additionally, we enforce convexity of the neural network by (1) forcing all the weights of the neural network to be non-negative, and (2) using a non-decreasing convex activation function~\cite{sivaprasad2021curious}. Non-negative weights are achieved by clamping the weights after every training epoch. This improves with convergence and reduces numerical errors for the Newton's method-based iterative solver~\cite{boyd2004convex, Argyros2016}.

\subsection{KNODE Training}
To train the neural network, we employ the strategy in~\cite{Jiahao2021Knowledgebased}, where model predictions are made with each point along the spatial trajectory as the initial condition. The predictions are then compared with the ground truth trajectory to compute the loss. The ground truth, i.e. training data, is generated with the ``correct'' system. Overall, the loss function is given by
\begin{multline}
    \label{eqn: loss_function}
    L(\boldsymbol{\theta}) = \frac{1}{|\mathbb{S}|(|\mathbb{T_{\mathcal{O}}}|-1)}\sum_{s \in \mathbb{S}}\sum_{\substack{t \in \mathbb{T_{\mathcal{O}}}\\
    t \neq t_0}}\Big(\|\Tilde{\bfy}(s, t) - \bfy(s, t)\|_2^2 + \\
     \|\Tilde{\bfz}(s, t) - \bfz(s, t)\|_2^2\Big) + \mathcal{R}(\boldsymbol{\theta}),
\end{multline}
where $\mathbb{S}$ is a subset of points in spatial discretization, $\mathbb{T_{\mathcal{O}}}$ is the set of the temporal discretizations, $\mathcal{R}$ is the regularization term on the neural network weights, and $\Tilde{\bfy}(s, t)$ is obtained through the spatial numerical integration given by
\begin{equation}
    \Tilde{\bfy}(s_n, t) = \bfy(s_{n-1}, t) + \int^{s_n}_{s_{n-1}} \Tilde{\bfy}_s ds,
\end{equation}
where the initial conditions $\bfy(s_{n-1}, t)$ of this integration is directly taken from the observation data, and $\Tilde{\bfz}(s, t)$ can be computed using the second equation in \eqref{eq:semi-discretized general form}.

\section{Experiments and Results}
\subsection{Tendon-driven Continuum Robot}
\begin{table}[ht]
\caption{Notations and definitions of continuum robot dynamics}
\centering
\begin{tabular}{|l|l|l|}
\hline
\textbf{Symbol}                & \textbf{Units}                   & \textbf{Definition}                                                                                                         \\ \hline
$s$                   & m                       & Reference arclength                                                                                                \\
$t$                   & s                       & Time                                                                                                               \\
$\boldsymbol{p}$      & m                       & Global position in Cartesian coordinates                                                                           \\
$R$                   & none                    & Rotation matrix of material orientation                                                                            \\
$\boldsymbol{h}$      & none                    & Quaternion for the material orientation                                                                            \\
$\boldsymbol{n}$      & N                       & Internal force in the global frame                                                                                 \\
$\boldsymbol{m}$      & Nm                      & Internal moment in the global frame                                                                                \\
$\boldsymbol{f}$      & N/m                     & Distributed force in the global frame                                                                              \\
$\boldsymbol{l}$      & Nm/m                    & Distributed moment in the global frame                                                                             \\
$\boldsymbol{v}$      & none                    & \begin{tabular}[c]{@{}l@{}}Rate of change of position with repsect \\ to arclength in the local frame\end{tabular} \\
$\boldsymbol{u}$      & 1/m                     & Curvature vector in the local frame                                                                                \\
$\boldsymbol{q}$      & m/s                     & Velocity in the local frame                                                                                        \\
$\boldsymbol{\omega}$ & 1/s                     & Angular velocity in the local frame                                                                                \\
$A$                   & m$^2$    & Cross-sectional area                                                                                               \\
$\rho$                & kg/m$^3$ & Material density                                                                                                   \\
$J$                   & m\textasciicircum{}4    & Second mass moment of intertia tensor                                                                              \\
$\boldsymbol{v}^*$    & none                    & Value of v when n = v\_t = 0                                                                                       \\
$\boldsymbol{u}^*$    & 1/m                     & Local curvature when m = u\_t = 0                                                                                  \\
$K_{se}$              & N                       & Stiffness matrix for shear and extension                                                                           \\
$K_{bt}$              & Nm$^2$   & Stiffness matrix for bending and twisting                                                                          \\
$B_{se}$              & Ns                      & Damping matrix for shear and extension                                                                             \\
$B_{bt}$              & Nm$^2$s  & Damping matrix for bending and twisting                                                                            \\
$C$                     & kg/m$^2$ & Square law drag coefficient matrix                                                                                 \\
$g$                     & m/s$^2$  & Gravitational acceleration vector     \\
$\boldsymbol{\tau}$                     & N  & Tensions in all tendons              \\
$\boldsymbol{\phi}$                     & rad  & Directions of all tendons at the root

\\\hline
               
\end{tabular}

\label{tbl: continuum symbols and notations}
\end{table}

We validate the proposed framework using a tendon-driven continuum robot. Derived from Cosserat rod models~\cite{rucker2011statics}, the semi-discretized dynamics of the robot is given by
\begin{equation}
    \begin{aligned}
        \boldsymbol{p}_s &= R\boldsymbol{v},\\
        R_s &= R \hat{\boldsymbol{u}},\\
        \boldsymbol{n}_s &= R[\rho A(\hat{\boldsymbol{\omega}}\boldsymbol{q} + \boldsymbol{q}_t) + C\boldsymbol{q}\odot |\boldsymbol{q}|] - \rho A\boldsymbol{g} - \boldsymbol{\tau}\cdot \boldsymbol{\phi},\\
        \boldsymbol{m}_s &= \rho R (\hat{\boldsymbol{\omega}}J\boldsymbol{\omega} + J \boldsymbol{\omega}_t) - \hat{\boldsymbol{p}}_s\boldsymbol{n} - \boldsymbol{l},\\
        \boldsymbol{q}_s &= \boldsymbol{v}_t - \hat{\boldsymbol{u}}\boldsymbol{q}+\hat{\boldsymbol{\omega}}\boldsymbol{v}, \\
        \boldsymbol{\omega}_s &= \boldsymbol{u}_t - \hat{\boldsymbol{u}}\boldsymbol{\omega},
    \end{aligned}
    \label{eqn: cosserat spatial}
\end{equation}
where $\hat{\cdot}$ is the mapping from $\mathbb{R}^3$ to $\mathfrak{se}(3)$. In this paper, the same notations are used as in~\cite{till2019real}. For clarity, we relist the notation definitions in Table \ref{tbl: continuum symbols and notations}. Detailed derivations of the robot dynamics can be found in~\cite{rucker2011statics, till2019real}. The temporal derivatives are defined using history terms as
\begin{equation}
\begin{aligned}
    \boldsymbol{v}_t = c_0\boldsymbol{v} + \overset{\text{h}}{\boldsymbol{v}},& \quad
    \boldsymbol{u}_t = c_0\boldsymbol{u} + \overset{\text{h}}{\boldsymbol{u}} \\
    \boldsymbol{q}_t = c_0\boldsymbol{q} + \overset{\text{h}}{\boldsymbol{q}}, & \quad
    \boldsymbol{\omega}_t = c_0\boldsymbol{\omega} + \overset{\text{h}}{\boldsymbol{\omega}}.
\end{aligned}
\end{equation}
In addition to the terms associated with both spatial and temporal derivatives, the terms only associated with temporal derivatives are defined by
\begin{equation}
    \begin{aligned}
        \boldsymbol{v} &= (K_{se} + c_0 B_{se})^{-1} (R^T\boldsymbol{n} + K_{se}\boldsymbol{v}^* - B_{se}\overset{\text{h}}{\boldsymbol{v}}) \\
        \boldsymbol{u} &= (K_{bt} + c_0 B_{bt})^{-1} (R^T\boldsymbol{m} + K_{bt}\boldsymbol{u}^* - B_{bt}\overset{\text{h}}{\boldsymbol{u}}). 
    \end{aligned}
    \label{eqn: v u calc}
\end{equation}
To write this dynamics  in terms of \eqref{eq:semi-discretized general form}, we have $\boldsymbol{y} = [\boldsymbol{p}, \boldsymbol{R}, \boldsymbol{n}, \boldsymbol{m}, \boldsymbol{q}, \boldsymbol{\omega}]$, and $\boldsymbol{z} = [\boldsymbol{u}, \boldsymbol{v}]$. In this work, we use BDF2 as the choice of implicit method for semi-discretization. This method is given by
\begin{equation}
    \bfy_t(t_n) = c_0 \bfy(t_n) + c_1\bfy(t_{n-1}) + c_2\bfy(t_{n-2}),
    \label{eqn: BDF2}
\end{equation}
where the coefficients $c_0 = 1.5/\delta t$, $c_1 = -2/\delta t$, and $c_2 = 0.5/\delta t$, with $\delta t$ being the time step size. 

The robot is controlled by adjusting the tension in each tendon. Each tendon is first set to a \textit{base tension} to keep it taut, and additional tension is applied as needed to create load differences that actuate the robot.

\subsection{Numerical Experiment Setup}

Firstly, we conduct numerical experiments to demonstrate the effectiveness of the KNODE-Cosserat framework. We will show that it can learn the true dynamics of the tendon-driven robot when given (1) an imperfect model of the robot used as the knowledge, and (2) training data, which is the trajectory of the \textit{true} robot. The training data is generated by simulating a \textit{true} robot using the implicit-shooting method. The \textit{true} robot's parameters are directly taken from a real-world robot, as listed in Table \ref{tab:variables2}. We create an imperfect model by modifying the parameters or removing certain functional terms from the true model. This imperfect model is then used as the knowledge.

The simulation uses a time step of $0.05s$, and the centerbone of the tendon-driven robot is discretized into 10 segments. Note that for this numerical experiment, the KNODE-Cosserat model dynamics differs slightly from equation \eqref{eq:knode_dynamics} in that the history terms $\overset{\text{h}}{\bfy}_s$ and $\overset{\text{h}}{\bfz}$ are not included as the inputs to the neural network. Since this numerical example is a less complex learning problem compared to the real-world experiment, we find that the model can learn the true dynamics with only $\bfy$, $\bfz$, and $\boldsymbol{\tau}$. Restricting the dimensionality of the input to the neural network can further improve the sample efficiency~\cite{migenda2021adaptive}.

\begin{figure}
\includegraphics[width=0.85\linewidth]{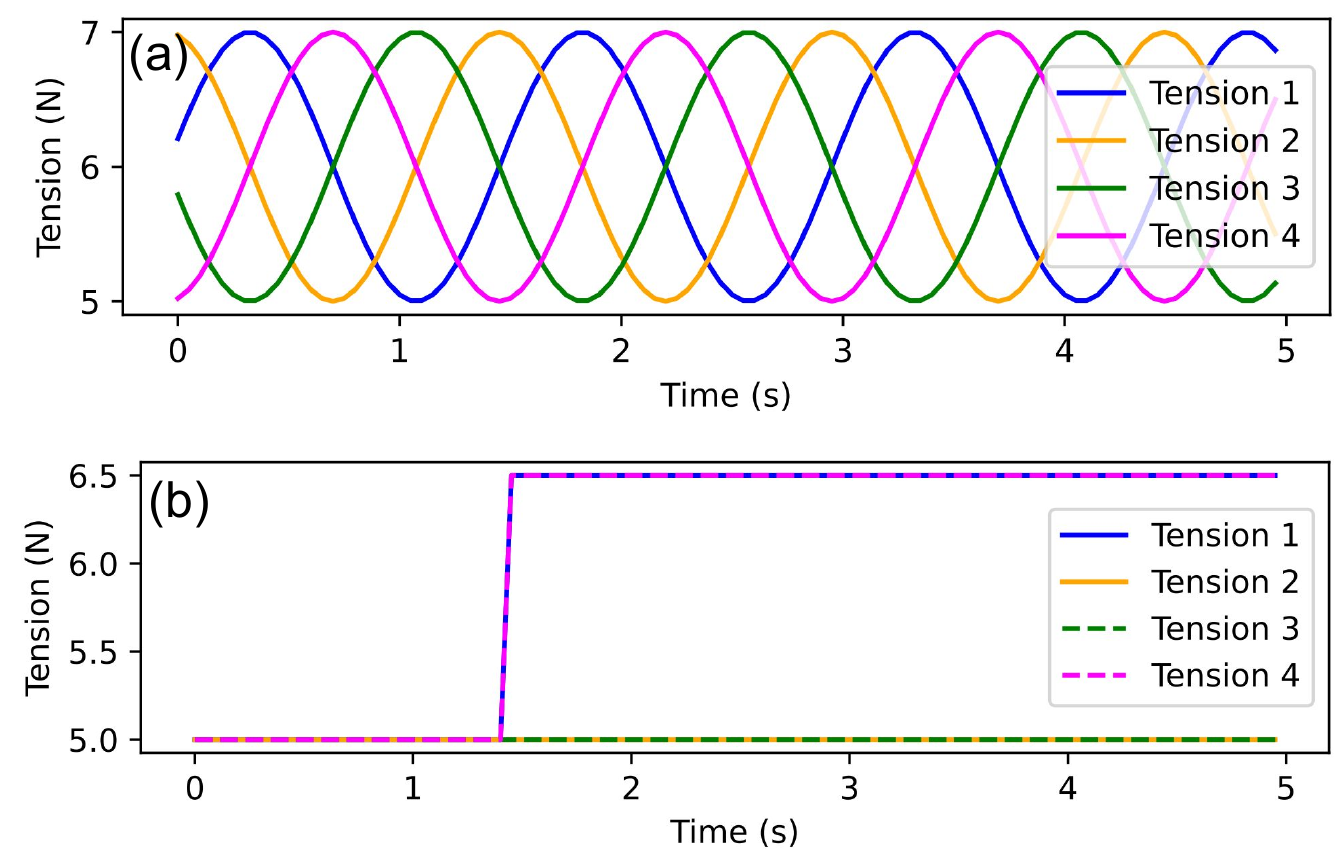}
    \caption{\textbf{Control inputs for simulation.} (a) Sine Controls. The sine waves have period $1.5s$. Tensions 1–4 are positioned counterclockwise about the rod. (b) Step Controls. Tensions 1–4 are positioned counterclockwise about the rod.}
    \label{fig:sineandstepinputs}
\end{figure}

We use a fully connected neural network with 1 hidden layer of 512 neurons, and Exponential Linear Unit (ELU) as the activation function. It is trained with Adam optimization using an initial learning rate of 0.01, $\beta$ values of 0.9 and 0.999, and no weight decay. The learning rate is scheduled using torch's \texttt{ReduceLROnPlateau} with patience 80 and factor 0.5. We train on 30 time steps (\textit{i.e.} 1.5 seconds) each from trajectories controlled with $0.5s$-period sine and with a $1s$-period sine similar to the control inputs shown in Fig. \ref{fig:sineandstepinputs}(a).

\begin{table}[ht]
    \caption{\textbf{Key Parameters for the True Tendon-driven Robot}}
    \centering
    \begin{tabular}{ | c | c | c | c | c | c | c | } 
      \hline
      L (m) & r (m) & $\rho$ (kg/m\textsuperscript{3}) & E (GPa) & B\textsubscript{bt} & B\textsubscript{se} & C \\ 
      \hline
      0.635 & 0.003175 & 1411.6751 & 2.757903 & 0.03 $I_3$ & 0 & 0.0001 \\
      \hline
    \end{tabular}
    \label{tab:variables2}
\end{table}

We use the following 4 modifications to the true parameters to create the imperfect model:

\begin{enumerate}
    \item Removing self-weight: We eliminate the $\rho Ag$ term in the dynamics (equation \ref{eqn: cosserat spatial}).
    \item Modifying length: We decrease rod length to $0.4m$.
    \item Modifying stiffness: We increase the Young's modulus to $10GPa$.
    \item Modifying stiffness \& length: a combination of the stiffness and length changes.
\end{enumerate}

The trained model is evaluated on two different sets of control inputs that were not part of the training data: (1) sinusoidal control with base tension of $6N$, amplitude of $1N$, period of $1.5s$, and phase offsets of $1/4$ (see Fig. \ref{fig:sineandstepinputs}(a)); and (2) step input by instantly increasing the tension in two adjacent tendons from $5N$ to $6.5N$ at time $t=1.5s$ (see Fig. \ref{fig:sineandstepinputs}(b)). For brevity, we refer to them as sine controls and step controls respectively. 

To evaluate the trained model, we simulate the trajectories from the true model, imperfect model, and the KNODE model for $100$ time steps, or $5s$, starting from an initial state of a stationary robot pointing directly upwards. We then compare how well the trajectories from the imperfect model and KNODE model match the true trajectory.

We use two metrics to evaluate how well the trajectories match: Euclidean distance between the positions of the tip of the rod in both trajectories after applying dynamic time warping (DTW), and the mean squared error (MSE) on the positions and orientations, expressed in Euler angles, of all segments of the discretized rod. In the subsequent sections, the "Reference" represents the trajectory of the true system, "Baseline" corresponds to the trajectory derived from the knowledge model, and "KNODE-Cosserat" refers to the trajectory generated by the trained model.

\subsection{Numerical Experiment Results}

\begin{figure}
    \centering
    \includegraphics[width=0.85\linewidth]{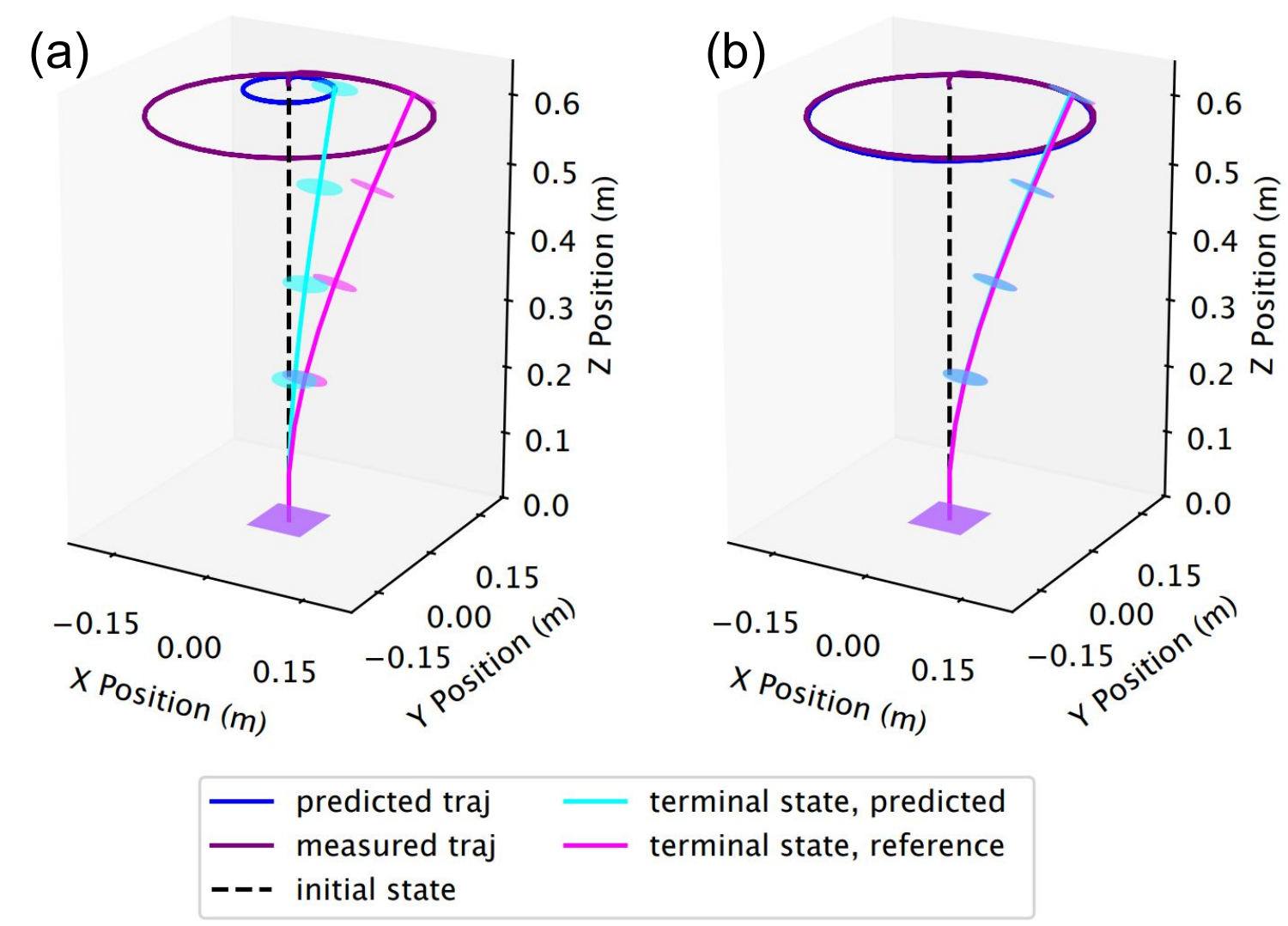}
    \caption{\textbf{Baseline and KNODE trajectories in simulation using imperfect stiffness.} (a) Trajectories from the imperfect model and the true model show different curvature under sine controls. (b) Trajectory generated using the KNODE model matches the true trajectory well.}
    \label{fig:simtraj}
\end{figure}

\begin{table}[ht]
\caption{\textbf{Simulation Results}}
\centering
\resizebox{0.9\linewidth}{!}{

\begin{tabular}{lcccccc}
\multicolumn{5}{l}{\textbf{Evaluation on Sine Controls}}                                                                                                                                                                                                                 \\ \hline
\multicolumn{1}{l|}{\multirow{2}{*}{\begin{tabular}[c]{@{}l@{}}Model\\ Imperfection\end{tabular}}} & \multicolumn{2}{c|}{Tip Position DTW}                                                  & \multicolumn{2}{c}{Pose MSE ($\times 10^3$)}              \\
\multicolumn{1}{l|}{}                                                                              & Baseline  & \multicolumn{1}{l|}{KNODE}                                                     & Baseline  & KNODE                                                     \\ \hline
\multicolumn{1}{l|}{Self-Weight}                                                                   & 1.75  & \multicolumn{1}{l|}{\begin{tabular}[c]{@{}c@{}}0.00\\ (-100.0\%)\end{tabular}} & 0.09  & \begin{tabular}[c]{@{}c@{}}0.00\\ (-100.0\%)\end{tabular} \\
\multicolumn{1}{l|}{Length}                                                                        & 40.16 & \multicolumn{1}{l|}{\begin{tabular}[c]{@{}c@{}}4.31\\ (-89.3\%)\end{tabular}}  & 16.46 & \begin{tabular}[c]{@{}c@{}}1.44\\ (-91.2\%)\end{tabular}  \\
\multicolumn{1}{l|}{Stiffness}                                                                     & 19.99 & \multicolumn{1}{l|}{\begin{tabular}[c]{@{}c@{}}0.94\\ (-95.3\%)\end{tabular}}  & 12.52 & \begin{tabular}[c]{@{}c@{}}0.03\\ (-99.8\%)\end{tabular}  \\
\multicolumn{1}{l|}{Length $\&$ Stiff}                                                             & 42.70 & \multicolumn{1}{l|}{\begin{tabular}[c]{@{}c@{}}5.07\\ (-88.1\%)\end{tabular}}  & 23.13 & \begin{tabular}[c]{@{}c@{}}1.31\\ (-94.4\%)\end{tabular}  \\
\multicolumn{5}{l}{\rule{0pt}{4ex} \textbf{Evaluation on Step Controls}}                                                                                                                                                                              \\ \hline
\multicolumn{1}{l|}{\multirow{2}{*}{\begin{tabular}[c]{@{}l@{}}Model\\ Imperfection\end{tabular}}} & \multicolumn{2}{c|}{Tip Position DTW}                                                  & \multicolumn{2}{c}{Pose MSE ($\times 10^3$)}              \\
\multicolumn{1}{l|}{}                                                                              & Baseline  & \multicolumn{1}{l|}{KNODE}                                                     & Baseline  & KNODE                                                     \\ \hline
\multicolumn{1}{l|}{Self-Weight}                                                                   & 1.30  & \multicolumn{1}{l|}{\begin{tabular}[c]{@{}c@{}}0.00\\ (-100.0\%)\end{tabular}} & 0.23  & \begin{tabular}[c]{@{}c@{}}0.00\\ (-100.0\%)\end{tabular} \\
\multicolumn{1}{l|}{Length}                                                                        & 32.88  & \multicolumn{1}{l|}{\begin{tabular}[c]{@{}c@{}}11.54\\ (-64.9\%)\end{tabular}} & 14.90 & \begin{tabular}[c]{@{}c@{}}7.71\\ (-48.3\%)\end{tabular} \\
\multicolumn{1}{l|}{Stiffness}                                                                     & 13.92  & \multicolumn{1}{l|}{\begin{tabular}[c]{@{}c@{}}2.44\\ (-82.5\%)\end{tabular}}  & 11.15 & \begin{tabular}[c]{@{}c@{}}0.19\\ (-98.3\%)\end{tabular}  \\
\multicolumn{1}{l|}{Length $\&$ Stiff}                                                             & 34.57 & \multicolumn{1}{l|}{\begin{tabular}[c]{@{}c@{}}10.95\\ (-68.3\%)\end{tabular}} & 21.44 & \begin{tabular}[c]{@{}c@{}}5.94\\ (-72.3\%)\end{tabular}
\end{tabular}
}
\label{tab:simresults}
\end{table}

\begin{figure}
    \centering
    \includegraphics[width=0.85\linewidth]{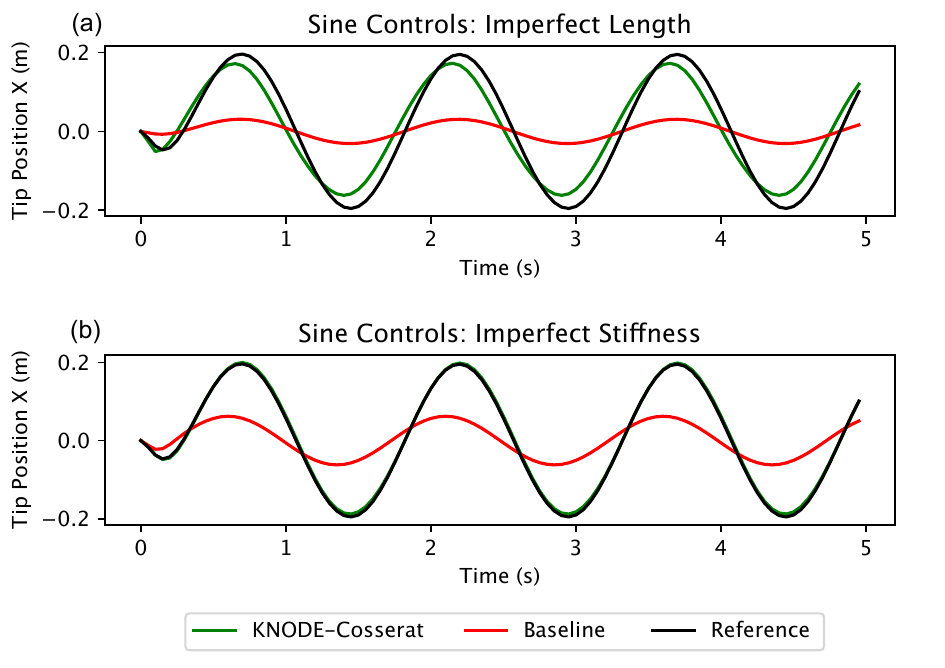}
    \caption{\textbf{Example tip trajectories evaluated on sine controls in simulation.} (a) The imperfect model has shorter length. (b) The imperfect model has stiffer centerbone.}
    \label{fig:simtrajsinegraph}
\end{figure}

Table \ref{tab:simresults} shows the simluation results on the performance of imperfect model (BASE) and KNODE-Cosserat model (KNODE) with respect to the reference robot. It can be seen that our KNODE-Cosserat framework reduces DTW distance by at least 64.9\% and MSE by at least 48.3\% compared to the imperfect model across all experiments. Qualitatively, Fig. \ref{fig:simtraj} shows how the KNODE model improves the trajectory accuracy when the knowledge has higher centerbone stiffness. While the imperfect model's trajectory with Sine Control has a much smaller radius compared to the true trajectory, KNODE's trajectory matches the true trajectory well. Furthermore, it can be seen from Fig. \ref{fig:simtrajsinegraph} that KNODE-Cosserat can correct for both the differences in frequency and amplitude in the trajectory. 

\begin{figure}
    \centering
    \includegraphics[width=0.85\linewidth]{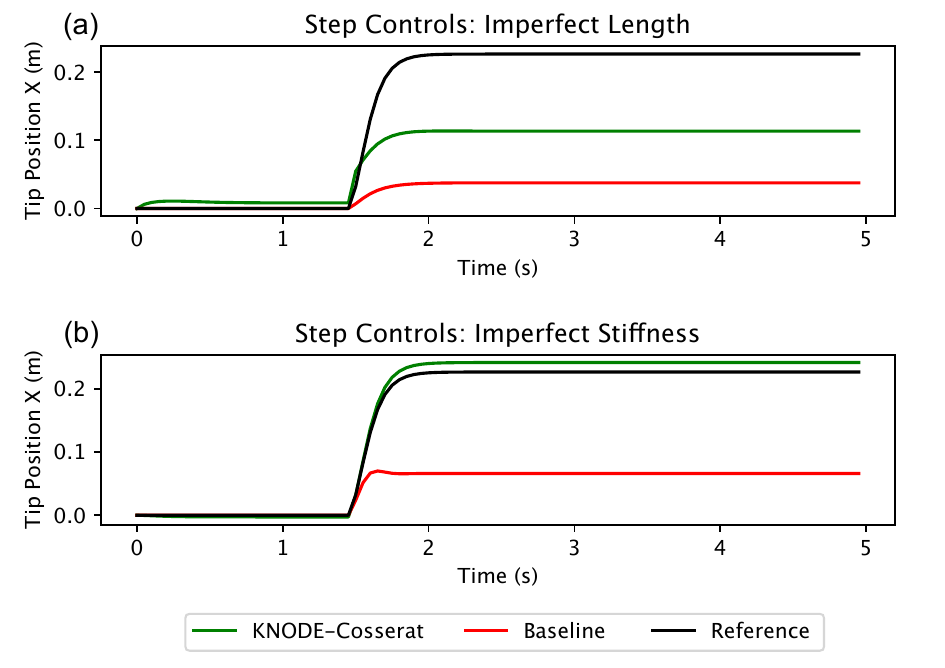}
    \caption{\textbf{Example tip trajectories evaluated on step controls in simulation.} (a) The imperfect model has shorter length. (b) The imperfect model has stiffer centerbone.}
    \label{fig:simtrajstepgraph}
\end{figure}

Figure \ref{fig:simtrajstepgraph} shows the step responses of the three rods. Because the model is trained on only sine trajectories, the trajectory accuracy improvement on the step controls is less than that of sine controls. Nevertheless, the step response from the KNODE model is closer in $x$ position to the true trajectory compared to the imperfect model.


\subsection{Real-world Experiments Setup}

\begin{figure}
    \centering
    \includegraphics[width=0.85\linewidth]{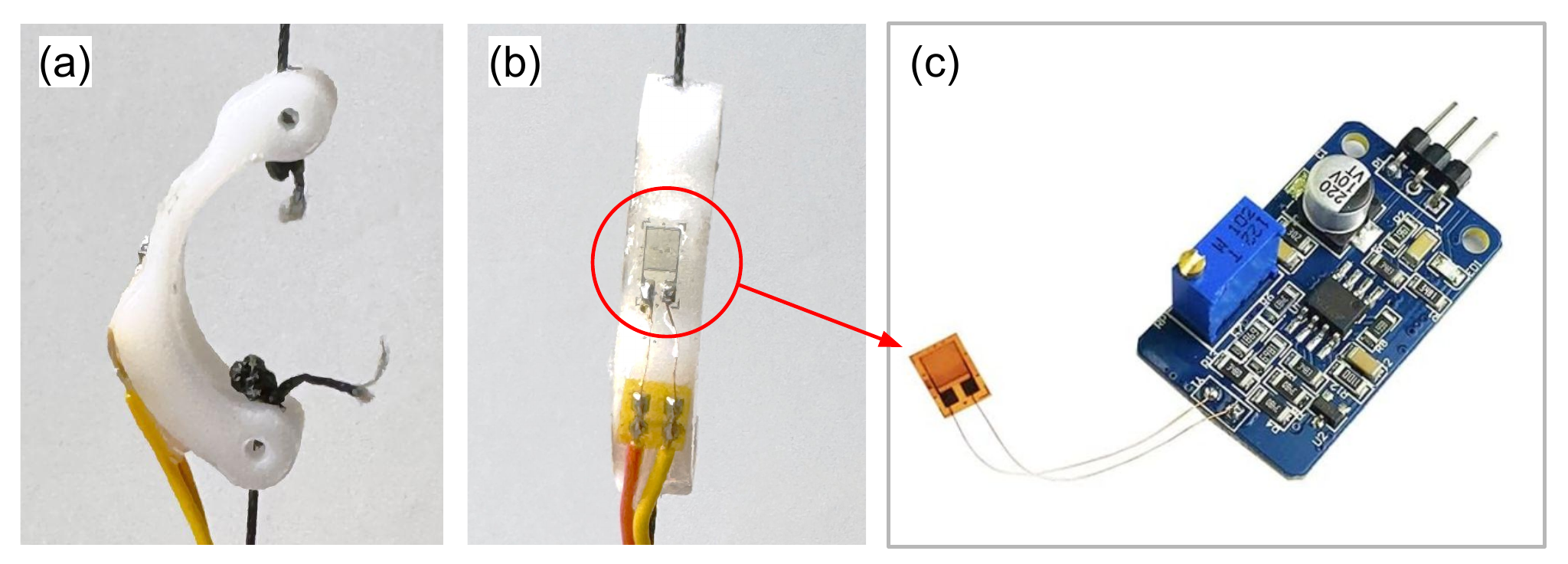}
    \caption{\textbf{Custom Load cell.} (a) Side view showing the curved design of the load cell. (b) Front view showing the strain gauge glued at the center of the body and wires secured at the bottom. (c) The amplifier module used to amplify the strain gauge signals.}
    \label{fig:custom load cell}
\end{figure}

For the real-world demonstration of our proposed algorithm, we will show that KNODE-Cosserat can effectively learn the dynamics of a real robot using trajectories generated by the robot and an imperfect physics model as prior knowledge. The experiment is structured as follows: first, we generate trajectories using various tension commands based on the knowledge model. Next, we replicate these trajectories with the real robot using the same commands. We then compare the trajectories by calculating DTW and MSE errors, as previously described. Following this, we train the KNODE-Cosserat model using the data collected from the real robot. Finally, we evaluate the trajectories produced by the KNODE-Cosserat model, demonstrating improvements in accuracy.


We custom-built a low-cost tendon-driven continuum robot, as shown in Fig. \ref{fig: custom robot annotated}, with open-source hardware and software available on the \href{https://github.com/hsiehScalAR/KNODE-Cosserat}{KNODE-Cosserat GitHub repository}. The robot has four equally spaced ribs rigidly attached to a 6.35mm Delrin® Acetal Plastic centerbone, with tendons running through each rib at 90° intervals. These tendons are controlled by reel mechanisms driven by four Seeed Studio JGA25‐371 Geared DC Motors (Fig. \ref{fig: custom robot annotated}(d)), allowing tension adjustment. The system is controlled by an Arduino Mega 2560 Rev3 and two dual H-bridge XY-160D motor controllers.

We measure tendon tension using a low-cost custom sensor, consisting of a Delrin® Acetal Plastic deformable body and a Taidacent strain gauge (model: BF350-3AA), highlighted in red in Fig. \ref{fig:custom
load cell}(b). The strain gauge includes an Amplifier Module shown in Fig. \ref{fig:custom load cell}(c). These custom load cells are calibrated against off-the-shelf models and complete the closed loop for the motor-driven reel mechanisms. A PID controller is used to track the desired tendon tension. Unlike our compact design, existing tendon-driven platforms rely on bulkier, more expensive linear actuators with built-in sensors~\cite{9762144}.

\begin{figure}
    \centering
    \includegraphics[width=0.85\linewidth]{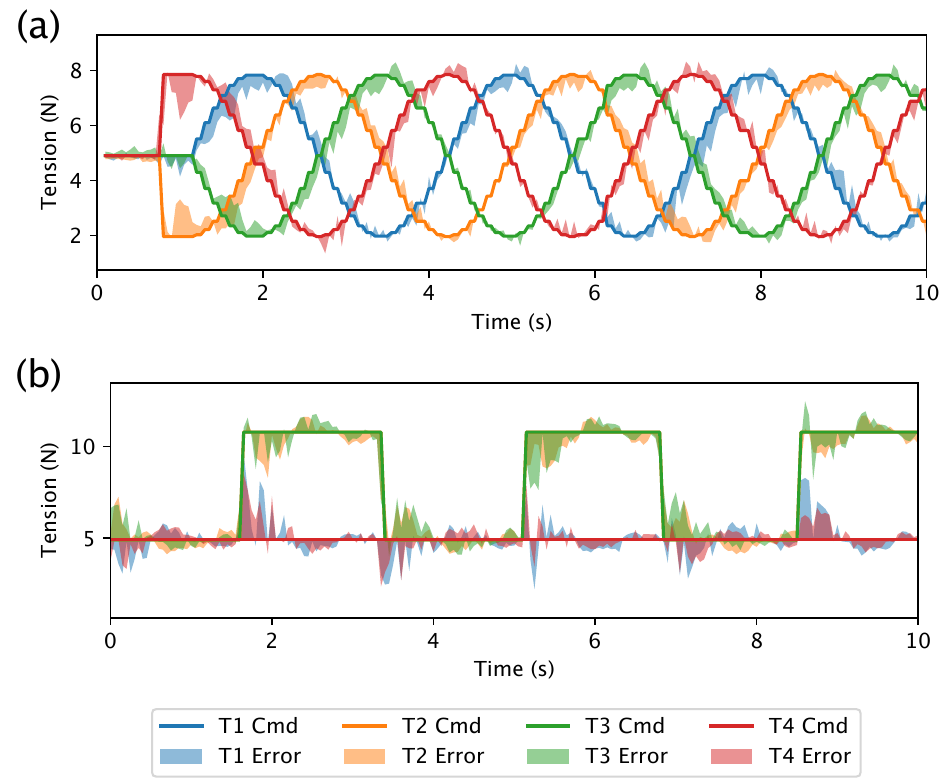}
    \caption{\textbf{Tension commands and the error between commanded and measured tension from the sensors} for (a) the Sine 3.0s controls and (b) the Step 10.8 N controls. The measured tension is used for training and evaluation.}
    \label{fig:realcontrols}
\end{figure}

A ROS node sends tension commands to the robot via a serial port (example in Fig. \ref{fig:realcontrols}). To validate control consistency, we repeat each control type 15 times and collect data using the Vicon motion capture system. The position and orientation at five key points along the robot are tracked and recorded. Since the data provides only a partial state, additional state estimation is required to work with the proposed method.

For state estimation, we first interpolated the position and orientation measurements through cubic splines. Then linear and angular velocities $\boldsymbol{q}$ and $\boldsymbol{\omega}$ are estimated using numerical differentiation on the positions and orientations. To estimate $\boldsymbol{n}$ and $\boldsymbol{m}$, we need to numerically integrate backwards from the tip of the robot to its root. Since we know that at the tip there is no internal force or moment, we have $\boldsymbol{n}(S, :) = 0$ and $\boldsymbol{m}(S, :) = 0$. We can then compute $\boldsymbol{n}$ and $\boldsymbol{m}$ iteratively for the entire rod by
\begin{equation}
    \begin{aligned}
        \boldsymbol{n}(s-1, :) &= \boldsymbol{n}(s, :) - \delta s \cdot \boldsymbol{n}_s, \\
        \boldsymbol{m}(s-1, :) &= \boldsymbol{m}(s, :) - \delta s \cdot \boldsymbol{m}_s,
    \end{aligned}
\end{equation}
where $\boldsymbol{n}_s$ and $\boldsymbol{m}_s$ are defined by \eqref{eqn: cosserat spatial}, and $\delta s$ is the size of spatial discretization. We can then estimate $\boldsymbol{v}$ and $\boldsymbol{u}$ using equation \eqref{eqn: v u calc}. While we only use this simplistic method for state estimation, we will show that KNODE-Cosserat can still improve the model accuracy.

The neural network used for real-world experiments has the same architecture as the simulation network, but is trained using a weight decay of 0.1. We also add noise with magnitude 0.01 and distributed over the standard normal to the trajectories in the training set. In dynamics learning this is known as \textit{stabilization noise}, which helps to stabilize training and improve convergence~\cite{OttRescomp}.

Note that this is a more challenging learning task than the numerical experiments. The ``true'' system is the physical hardware, and the differences between it and the knowledge could be from many different sources. For example, assumptions such as frictionless contact between the tendons and ribs, uniform load along the centerbone, isotropic material, and so on can all contribute to the difference between the true model and the imperfect model.

For evaluating performance on real-world data, we use the same lengths, parameters, metrics, and setup as the simulation.  The KNODE model is trained on two trajectory datasets:

\begin{itemize}
    \item Dataset A: Trajectories generated from sine controls with period $1s$ and period $3s$. Both have base tension of $4.9 N$ and amplitude of $2.9N$.
    \item Dataset B: The same two sine trajectories in dataset A, plus a trajectory using controls sampled from a uniform random distribution in the range from $4.9N$ to $11.8N$.
\end{itemize}

Example tension commands are shown in Fig. \ref{fig:realcontrols}.

\subsection{Real-world Experiments Results}

\begin{figure}[ht]
    \centering
    \includegraphics[width=0.85\linewidth]{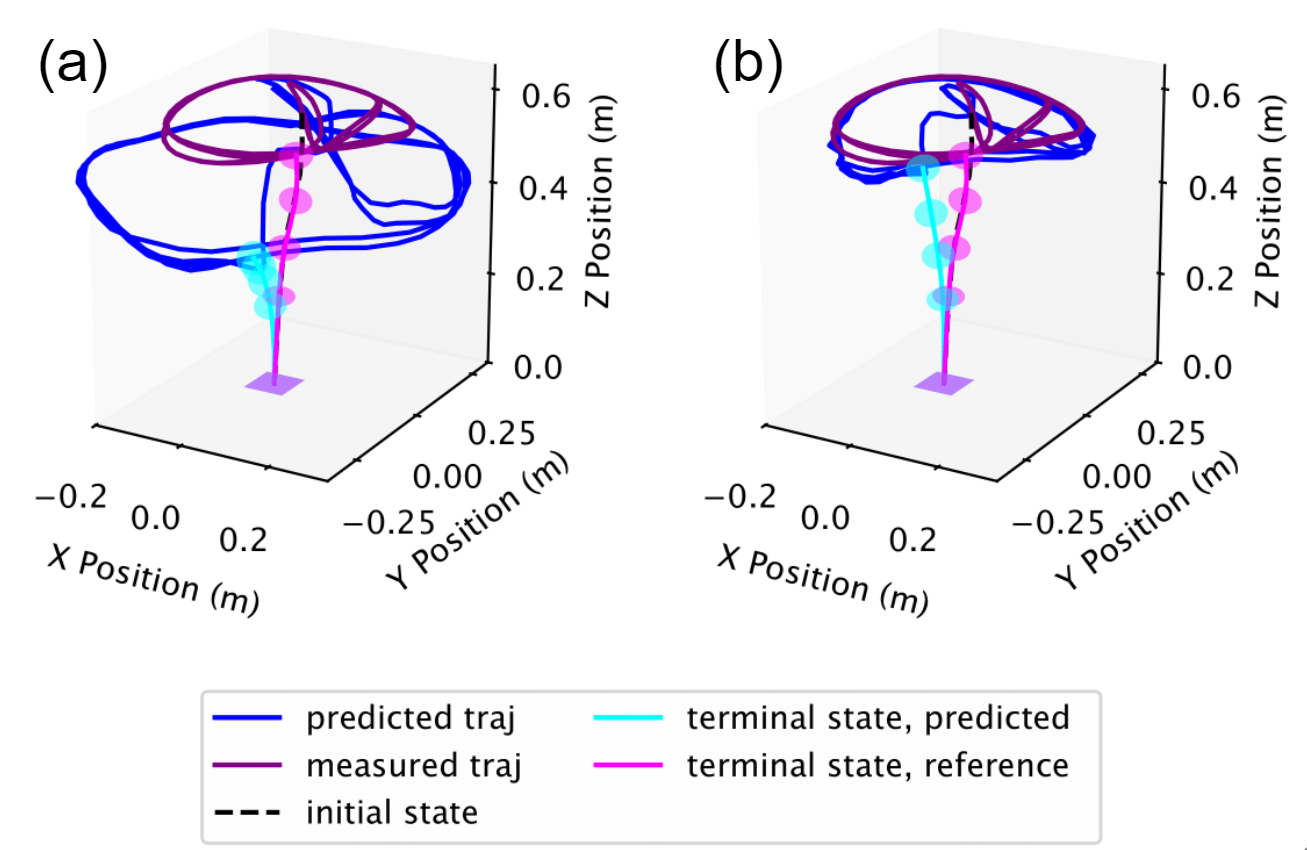}
    \caption{\textbf{The real-world baseline and knode trajectories for 2s-period sine}. Both are evaluated for 10s, and the knode trajectory is trained on dataset B (one random and two sine trajectories). (a) Trajectories from the imperfect model and the true model under sine controls show large discrepancies. (b) Trajectory generated using the KNODE model matches the true trajectory well.}
    \label{fig:real qualitative}
\end{figure}

\begin{figure}
    \centering    \includegraphics[width=0.85\linewidth]{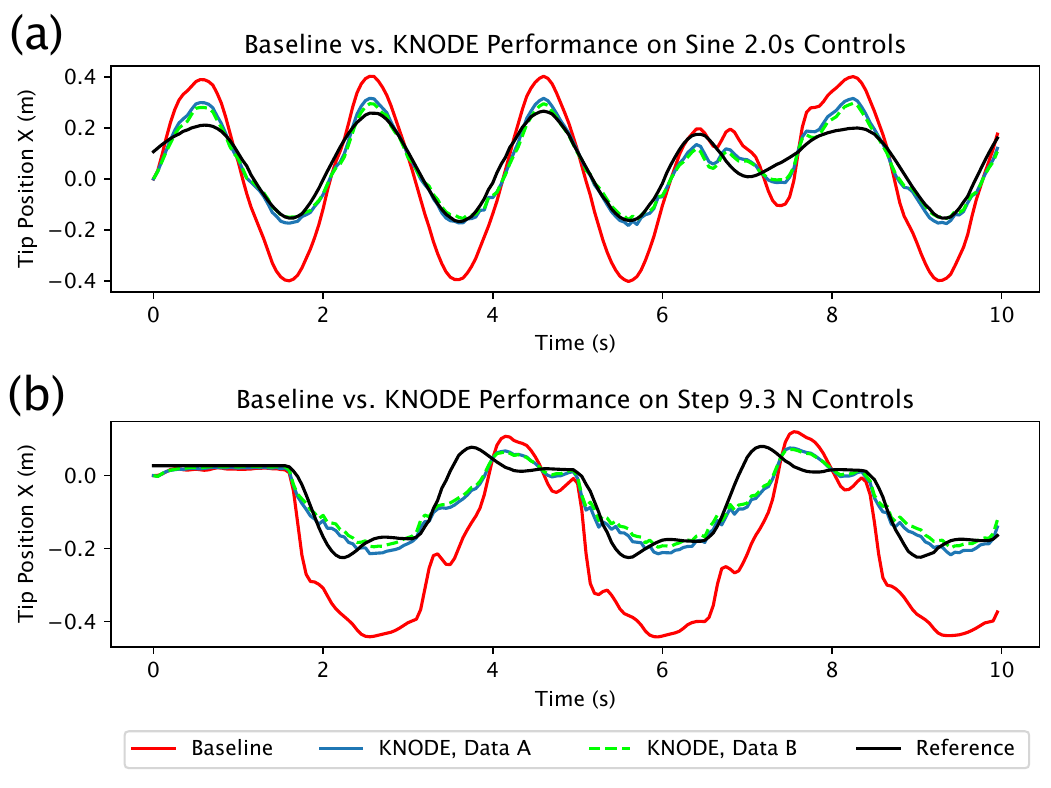}
    \caption{\textbf{Example tip trajectories in X with real-world robot.} (a) On Sine Controls. (b) On Step Controls.}
    \label{fig:xgraphs-real}
\end{figure}

Table \ref{tab:realresults} lists the results of the real-world experiment. It can be observed that both metrics show improvement from the baseline, with an average improvement of $58.7\%$. From Fig. \ref{fig:real qualitative} and Fig. \ref{fig:xgraphs-real}, it can also be observed that qualitatively KNODE-Cosserat improves the model and generates a trajectory that matches the true trajectory of the real robot. Notably, we find that for almost all cases, the model trained with the additional random trajectory has lower error.

\begin{figure}[H]
    \centering    \includegraphics[width=0.9\linewidth]{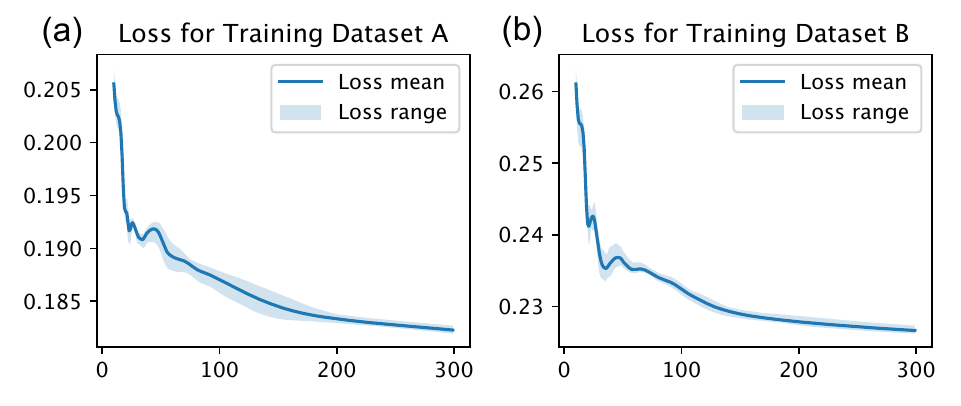}
    \caption{\textbf{Training loss curves for real-world robot.}}
    \label{fig:real loss}
\end{figure}

We also repeated the experiments using five different random seeds for initializing the neural network for both the numerical and real-world experiments. We observe consistent convergence and performance improvement across all instances. The training loss curves for the real-world experiments are shown in Fig. \ref{fig:real loss}.

\begin{table}[h]
\caption{\textbf{Real-world Experiment Results}}
\centering
\resizebox{\linewidth}{!}{
\begin{tabular}{l|lcc|ccc}
\multirow{2}{*}{\begin{tabular}[c]{@{}l@{}}Controls\end{tabular}} & \multicolumn{3}{l|}{Tip Position DTW} & \multicolumn{3}{l}{Pos $\&$ Rot MSE ($\times 10^3$)} \\
 & Baseline & \multicolumn{1}{l}{Data A} & \multicolumn{1}{l|}{Data B} & \multicolumn{1}{l}{Baseline} & \multicolumn{1}{l}{Data A} & \multicolumn{1}{l}{Data B} \\ 
\hline
Sine 0.5s & \multicolumn{1}{c}{41.46} & \begin{tabular}[c]{@{}c@{}}21.10\\(-49.1\%)\end{tabular} & \begin{tabular}[c]{@{}c@{}}18.50\\(-55.4\%)\end{tabular} & 37.55 & \begin{tabular}[c]{@{}c@{}}16.27\\(-56.7\%)\end{tabular} & \begin{tabular}[c]{@{}c@{}}12.38\\(-67.0\%)\end{tabular} \\
Sine 0.75s & 41.84 & \begin{tabular}[c]{@{}c@{}}21.14\\(-49.5\%)\end{tabular} & \begin{tabular}[c]{@{}c@{}}19.58\\(-53.2\%)\end{tabular} & 44.00 & \begin{tabular}[c]{@{}c@{}}18.16\\(-58.7\%)\end{tabular} & \begin{tabular}[c]{@{}c@{}}14.23\\(-67.7\%)\end{tabular} \\
Sine 1.0s & 28.62 & \begin{tabular}[c]{@{}c@{}}23.02\\(-19.6\%)\end{tabular} & \begin{tabular}[c]{@{}c@{}}23.08\\(-19.4\%)\end{tabular} & 41.31 & \begin{tabular}[c]{@{}c@{}}20.57\\(-50.2\%)\end{tabular} & \begin{tabular}[c]{@{}c@{}}17.17\\(-58.4\%)\end{tabular} \\
Sine 2.0s & 67.57 & \begin{tabular}[c]{@{}c@{}}17.46\\(-74.2\%)\end{tabular} & \begin{tabular}[c]{@{}c@{}}13.84\\(-79.5\%)\end{tabular} & 49.36 & \begin{tabular}[c]{@{}c@{}}12.69\\(-74.3\%)\end{tabular} & \begin{tabular}[c]{@{}c@{}}9.25\\(-81.3\%)\end{tabular} \\
Sine 3.0s & 84.79 & \begin{tabular}[c]{@{}c@{}}31.00\\(-63.4\%)\end{tabular} & \begin{tabular}[c]{@{}c@{}}25.78\\(-69.6\%)\end{tabular} & 65.05 & \begin{tabular}[c]{@{}c@{}}15.87\\(-75.6\%)\end{tabular} & \begin{tabular}[c]{@{}c@{}}11.26\\(-82.7\%)\end{tabular} \\
Step 7.8 N & 35.21 & \begin{tabular}[c]{@{}c@{}}14.18\\(-59.7\%)\end{tabular} & \begin{tabular}[c]{@{}c@{}}11.41\\(-67.6\%)\end{tabular} & 28.15 & \begin{tabular}[c]{@{}c@{}}8.33\\(-70.4\%)\end{tabular} & \begin{tabular}[c]{@{}c@{}}5.45\\(-80.6\%)\end{tabular} \\
Step 9.3 N & 40.41 & \begin{tabular}[c]{@{}c@{}}15.42\\(-61.8\%)\end{tabular} & \begin{tabular}[c]{@{}c@{}}12.45\\(-69.2\%)\end{tabular} & 36.35 & \begin{tabular}[c]{@{}c@{}}13.54\\(-62.7\%)\end{tabular} & \begin{tabular}[c]{@{}c@{}}9.32\\(-74.4\%)\end{tabular} \\
Step 10.8 N & 41.80 & \begin{tabular}[c]{@{}c@{}}18.74\\(-55.2\%)\end{tabular} & \begin{tabular}[c]{@{}c@{}}16.68\\(-60.1\%)\end{tabular} & 79.13 & \begin{tabular}[c]{@{}c@{}}36.56\\(-53.8\%)\end{tabular} & \begin{tabular}[c]{@{}c@{}}28.21\\(-64.3\%)\end{tabular} \\
Step 12.3 N & 39.63 & \begin{tabular}[c]{@{}c@{}}22.39\\(-43.5\%)\end{tabular} & \begin{tabular}[c]{@{}c@{}}21.20\\(-46.5\%)\end{tabular} & 58.73 & \begin{tabular}[c]{@{}c@{}}30.14\\(-48.7\%)\end{tabular} & \begin{tabular}[c]{@{}c@{}}22.66\\(-61.4\%)\end{tabular} \\
Step 13.7 N & 35.05 & \begin{tabular}[c]{@{}c@{}}25.60\\(-27.0\%)\end{tabular} & \begin{tabular}[c]{@{}c@{}}25.11\\(-28.4\%)\end{tabular} & 62.01 & \begin{tabular}[c]{@{}c@{}}32.57\\(-47.5\%)\end{tabular} & \begin{tabular}[c]{@{}c@{}}24.48\\(-60.5\%)\end{tabular}
\end{tabular}
}
\label{tab:realresults}
\end{table}

\section{Conclusion and Future Work}
In this paper, we introduced KNODE-Cosserat, a data-driven modeling approach for Cosserat rod-based continuum robots. Our method combines first-principle physics models with artificial neural networks to improve the model accuracy. We validated its efficacy through both simulation and real-world experiments. On average, KNODE-Cosserat is shown to improve the model accuracy of Cosserat rod-based robot by $87.0\%$ in simulation and $58.7\%$ in real-world experiment, even for long-horizon trajectories where errors tend to compound. For future work, we will incorporate KNODE models in model-based control schemes, such as model predictive control, to achieve more accurate control of Cosserat rod-based robot. We also plan to explore other general classes of physics model for soft robot to incorporate KNODE.

\bibliographystyle{IEEEtran} 
\bibliography{IEEEabrv, refs}

\begin{thebibliography}{10}
\providecommand{\url}[1]{#1}
\csname url@samestyle\endcsname
\providecommand{\newblock}{\relax}
\providecommand{\bibinfo}[2]{#2}
\providecommand{\BIBentrySTDinterwordspacing}{\spaceskip=0pt\relax}
\providecommand{\BIBentryALTinterwordstretchfactor}{4}
\providecommand{\BIBentryALTinterwordspacing}{\spaceskip=\fontdimen2\font plus
\BIBentryALTinterwordstretchfactor\fontdimen3\font minus \fontdimen4\font\relax}
\providecommand{\BIBforeignlanguage}[2]{{%
\expandafter\ifx\csname l@#1\endcsname\relax
\typeout{** WARNING: IEEEtran.bst: No hyphenation pattern has been}%
\typeout{** loaded for the language `#1'. Using the pattern for}%
\typeout{** the default language instead.}%
\else
\language=\csname l@#1\endcsname
\fi
#2}}
\providecommand{\BIBdecl}{\relax}
\BIBdecl

\bibitem{Rus2015}
D.~Rus and M.~T. Tolley, ``Design, fabrication and control of soft robots,'' \emph{Nature}, vol. 521, no. 7553, pp. 467--475, 2015.

\bibitem{doi:10.1126/scirobotics.aah3690}
C.~Laschi, B.~Mazzolai, and M.~Cianchetti, ``Soft robotics: Technologies and systems pushing the boundaries of robot abilities,'' \emph{Science Robotics}, vol.~1, no.~1, p. eaah3690, 2016.

\bibitem{trivedi}
D.~Trivedi, C.~D. Rahn, W.~M. Kier, and I.~D. Walker, ``Soft robotics: Biological inspiration, state of the art, and future research,'' \emph{Applied Bionics and Biomechanics}, vol.~5, no.~3, p. 520417, 2008.

\bibitem{till2019real}
J.~Till, V.~Aloi, and C.~Rucker, ``Real-time dynamics of soft and continuum robots based on cosserat rod models,'' \emph{The International Journal of Robotics Research}, vol.~38, no.~6, pp. 723--746, 2019.

\bibitem{rucker2011statics}
D.~C. Rucker and R.~J. Webster~III, ``Statics and dynamics of continuum robots with general tendon routing and external loading,'' \emph{IEEE Transactions on Robotics}, vol.~27, no.~6, pp. 1033--1044, 2011.

\bibitem{QinReport}
L.~Qin, H.~Peng, X.~Huang, M.~Liu, and W.~Huang, ``Modeling and simulation of dynamics in soft robotics: a review of numerical approaches,'' \emph{Current Robotics Reports}, vol.~5, pp. 1--13, 08 2023.

\bibitem{Bauchau2009}
O.~A. Bauchau and J.~I. Craig, \emph{Euler-Bernoulli beam theory}.\hskip 1em plus 0.5em minus 0.4em\relax Dordrecht: Springer Netherlands, 2009, pp. 173--221.

\bibitem{MbakopInverseDynamics}
S.~Mbakop, G.~Tagne, M.-H. Frouin, M.~Achille, and R.~Merzouki, ``Inverse dynamics model-based shape control of soft continuum finger robot using parametric curve,'' \emph{IEEE Robotics and Automation Letters}, vol.~6, pp. 8053--8060, 10 2021.

\bibitem{Webster2010DesignAK}
R.~J. Webster and B.~A. Jones, ``Design and kinematic modeling of constant curvature continuum robots: A review,'' \emph{The International Journal of Robotics Research}, vol.~29, pp. 1661 -- 1683, 2010.

\bibitem{Bergou2008DiscreteER}
M.~Bergou, M.~Wardetzky, S.~Robinson, B.~Audoly, and E.~Grinspun, ``Discrete elastic rods,'' \emph{ACM SIGGRAPH 2008 papers}, 2008.

\bibitem{zhang:hal-01370347}
Z.~Zhang, J.~Dequidt, A.~Kruszewski, F.~Largilliere, and C.~Duriez, ``{Kinematic Modeling and Observer Based Control of Soft Robot using Real-Time Finite Element Method},'' in \emph{{IROS2016 - IEEE/RSJ International Conference on Intelligent Robots and Systems}}, Daejeon, South Korea, Oct. 2016.

\bibitem{Guo2020SimulationAF}
N.~Guo, Z.~Sun, X.~Wang, E.~H.~K. Yeung, M.~K.-T. To, X.~Li, and Y.~Hu, ``Simulation analysis for optimal design of pneumatic bellow actuators for soft-robotic glove,'' \emph{Biocybernetics and Biomedical Engineering}, vol.~40, pp. 1359--1368, 2020.

\bibitem{ChenClimb}
G.~Chen, T.~Lin, G.~Lodewijks, and A.~Ji, ``Design of an active flexible spine for wall climbing robot using pneumatic soft actuators,'' \emph{Journal of Bionic Engineering}, vol.~20, 09 2022.

\bibitem{Goury2018FastGA}
O.~Goury and C.~Duriez, ``Fast, generic, and reliable control and simulation of soft robots using model order reduction,'' \emph{IEEE Transactions on Robotics}, vol.~34, pp. 1565--1576, 2018.

\bibitem{katzschmann2019dynamically}
R.~K. Katzschmann, M.~Thieffry, O.~Goury, A.~Kruszewski, T.-M. Guerra, C.~Duriez, and D.~Rus, ``Dynamically closed-loop controlled soft robotic arm using a reduced order finite element model with state observer,'' in \emph{2019 2nd IEEE international conference on soft robotics (RoboSoft)}.\hskip 1em plus 0.5em minus 0.4em\relax IEEE, 2019, pp. 717--724.

\bibitem{li2022equivalent}
S.~Li, A.~Kruszewski, T.-M. Guerra, and A.-T. Nguyen, ``Equivalent-input-disturbance-based dynamic tracking control for soft robots via reduced-order finite-element models,'' \emph{IEEE/ASME Transactions on Mechatronics}, vol.~27, no.~5, pp. 4078--4089, 2022.

\bibitem{RAISSI2019686}
M.~Raissi, P.~Perdikaris, and G.~Karniadakis, ``Physics-informed neural networks: A deep learning framework for solving forward and inverse problems involving nonlinear partial differential equations,'' \emph{Journal of Computational Physics}, vol. 378, pp. 686--707, 2019.

\bibitem{zhu2019physics}
Y.~Zhu, N.~Zabaras, P.-S. Koutsourelakis, and P.~Perdikaris, ``Physics-constrained deep learning for high-dimensional surrogate modeling and uncertainty quantification without labeled data,'' \emph{Journal of Computational Physics}, vol. 394, pp. 56--81, 2019.

\bibitem{Shi2023}
L.~Shi, Z.~Liu, and K.~Karydis, ``Koopman operators for modeling and control of soft robotics,'' \emph{Current Robotics Reports}, vol.~4, no.~2, pp. 23--31, 2023.

\bibitem{shi2024koopman}
L.~Shi, M.~Haseli, G.~Mamakoukas, D.~Bruder, I.~Abraham, T.~Murphey, J.~Cortes, and K.~Karydis, ``Koopman operators in robot learning,'' \emph{arXiv preprint arXiv:2408.04200}, 2024.

\bibitem{9369003}
N.~Naughton, J.~Sun, A.~Tekinalp, T.~Parthasarathy, G.~Chowdhary, and M.~Gazzola, ``Elastica: A compliant mechanics environment for soft robotic control,'' \emph{IEEE Robotics and Automation Letters}, vol.~6, no.~2, pp. 3389--3396, 2021.

\bibitem{9561145}
G.~Li, J.~Shintake, and M.~Hayashibe, ``Deep reinforcement learning framework for underwater locomotion of soft robot,'' in \emph{2021 IEEE International Conference on Robotics and Automation (ICRA)}, 2021, pp. 12\,033--12\,039.

\bibitem{Truby2020DistributedPO}
R.~L. Truby, C.~D. Santina, and D.~Rus, ``Distributed proprioception of 3d configuration in soft, sensorized robots via deep learning,'' \emph{IEEE Robotics and Automation Letters}, vol.~5, pp. 3299--3306, 2020.

\bibitem{fang2022efficient}
G.~Fang, Y.~Tian, Z.-X. Yang, J.~M. Geraedts, and C.~C. Wang, ``Efficient jacobian-based inverse kinematics with sim-to-real transfer of soft robots by learning,'' \emph{IEEE/ASME Transactions on Mechatronics}, vol.~27, no.~6, pp. 5296--5306, 2022.

\bibitem{Jiahao2021Knowledgebased}
T.~Z. Jiahao, M.~A. Hsieh, and E.~Forgoston, ``Knowledge-based learning of nonlinear dynamics and chaos,'' \emph{Chaos}, vol.~31, no.~11, p. 111101, 2021.

\bibitem{jiahao2022online}
T.~Z. Jiahao, K.~Y. Chee, and M.~A. Hsieh, ``Online dynamics learning for predictive control with an application to aerial robots,'' in \emph{6th Annual Conference on Robot Learning}, 2022.

\bibitem{chee2022knode}
K.~Y. Chee, T.~Z. Jiahao, and M.~A. Hsieh, ``Knode-mpc: A knowledge-based data-driven predictive control framework for aerial robots,'' \emph{IEEE Robotics and Automation Letters}, vol.~7, no.~2, pp. 2819--2826, 2022.

\bibitem{9811997}
T.~Z. Jiahao, L.~Pan, and M.~A. Hsieh, ``Learning to swarm with knowledge-based neural ordinary differential equations,'' in \emph{2022 International Conference on Robotics and Automation (ICRA)}, 2022, pp. 6912--6918.

\bibitem{xun2024cosserat}
L.~Xun, G.~Zheng, and A.~Kruszewski, ``Cosserat-rod based dynamic modeling of soft slender robot interacting with environment,'' \emph{IEEE Transactions on Robotics}, 2024.

\bibitem{wang2024cosserat}
X.~Wang and N.~Rojas, ``Cosserat rod modeling and validation for a soft continuum robot with self-controllable variable curvature,'' in \emph{2024 IEEE 7th International Conference on Soft Robotics (RoboSoft)}.\hskip 1em plus 0.5em minus 0.4em\relax IEEE, 2024, pp. 705--710.

\bibitem{197970}
T.~Na, ``Chapter 5 iterative methods—the shooting methods,'' in \emph{Computational Methods in Engineering Boundary Value Problems}, ser. Mathematics in Science and Engineering.\hskip 1em plus 0.5em minus 0.4em\relax Elsevier, 1979, vol. 145, pp. 70--92.

\bibitem{conf/nips/ChenRBD18}
T.~Q. Chen, Y.~Rubanova, J.~Bettencourt, and D.~Duvenaud, ``Neural ordinary differential equations.'' in \emph{NeurIPS}, 2018, pp. 6572--6583.

\bibitem{sivaprasad2021curious}
S.~Sivaprasad, A.~Singh, N.~Manwani, and V.~Gandhi, ``The curious case of convex neural networks,'' in \emph{Machine Learning and Knowledge Discovery in Databases. Research Track: European Conference, ECML PKDD 2021, Bilbao, Spain, September 13--17, 2021, Proceedings, Part I 21}.\hskip 1em plus 0.5em minus 0.4em\relax Springer, 2021, pp. 738--754.

\bibitem{boyd2004convex}
S.~Boyd and L.~Vandenberghe, \emph{Convex optimization}.\hskip 1em plus 0.5em minus 0.4em\relax Cambridge university press, 2004.

\bibitem{Argyros2016}
I.~K. Argyros and D.~Gonz{\'a}lez, \emph{Newton's Method for Convex Optimization}.\hskip 1em plus 0.5em minus 0.4em\relax Cham: Springer International Publishing, 2016, pp. 23--56.

\bibitem{migenda2021adaptive}
N.~Migenda, R.~M{\"o}ller, and W.~Schenck, ``Adaptive dimensionality reduction for neural network-based online principal component analysis,'' \emph{PloS one}, vol.~16, no.~3, p. e0248896, 2021.

\bibitem{9762144}
B.~Deutschmann, J.~Reinecke, and A.~Dietrich, ``Open source tendon-driven continuum mechanism: A platform for research in soft robotics,'' in \emph{2022 IEEE 5th International Conference on Soft Robotics (RoboSoft)}, 2022, pp. 54--61.

\bibitem{OttRescomp}
A.~Wikner, J.~Pathak, B.~Hunt, M.~Girvan, T.~Arcomano, I.~Szunyogh, A.~Pomerance, and E.~Ott, ``Combining machine learning with knowledge-based modeling for scalable forecasting and subgrid-scale closure of large, complex, spatiotemporal systems,'' \emph{Chaos}, vol.~30, p. 053111, 05 2020.

\end{thebibliography}

\newpage
\end{document}